\documentclass[runningheads]{llncs}
\usepackage{graphicx}
\usepackage{multirow}
\usepackage{subfigure}
\usepackage[colorinlistoftodos,prependcaption,textsize=tiny]{todonotes}
\begin{document}
\title{Sentiment Analysis from Images of Natural Disasters}
%
%
\author{Syed Zohaib Hassan \inst{1} \and
Kashif Ahmad\inst{2} \and
Ala Al-Fuqaha\inst{2} \and
Nicola Conci \inst{1}}
\authorrunning{S. Hassan et al.}
%
\institute{University of Trento, Trento Italy \and
Hamad Bin Khalifa University, Doha, Qatar \\
\email{syedzohaib.hassan@studenti.unitn.it} \\ \email{kahmad@hbku.edu.qa}, \email{aalfuqaha@hbku.edu.qa }, \email{nicola.conci@unitn.it}}
\maketitle              
\begin{abstract}
Social media have been widely exploited to detect and gather relevant information about opinions and events. However, the relevance of the information is very subjective and rather depends on the application and the end-users. In this article, we tackle a specific facet of social media data processing, namely the sentiment analysis of disaster-related images by considering people's opinions, attitudes, feelings and emotions. We analyze how visual sentiment analysis can improve the results for the end-users/beneficiaries in terms of mining information from social media.
We also identify the challenges and related applications, which could help defining a benchmark for future research efforts in visual sentiment analysis.

\end{abstract}
\section{Introduction}
Sudden and unexpected adverse events, such as floods and earthquakes, may not only damage the infrastructure but also have a significant impact on people's physical and mental health. In such events, an instant access to relevant information might help to identify and mitigate the damage. To this aim, information available on social networks can be utilized for the analysis of the potential impact of natural or man-made disasters on the environment and human lives \cite{ahmad2018social}. 

Social media outlets along with other sources of information, such as satellite imagery and Geographic Information Systems (GIS), have been widely exploited to provide a better coverage of natural and man-made disasters \cite{nogueira2018exploiting,ahmad2019automatic}. The majority of the approaches rely on computer vision and machine learning techniques to automatically detect disasters, collect, classify, and summarize relevant information. However, the interpretation of \textit{relevance} is very subjective and highly depends on the application framework and the end-users.  

In this article, we analyze the problem from a different perspective and focus in particular on sentiment analysis of disaster-related images. Specifically, we consider people's opinions, attitudes, feelings, and emotions toward the images related to the event by estimating the emotion/perceptual content evoked by a generic image \cite{constantin2019computational,gygli2013interestingness,machajdik2010affective}. We aim to explore and analyze how the visual sentiment analysis of such images can be utilized to provide more accurate description of adverse events, their evolution, and consequences. We believe that such analysis can serve as an effective tool to convey public sentiments around the world while reducing the bias of news organizations. This can lead to new beneficiaries beyond the general public (e.g., online news, humanitarian organizations, non-governmental organizations, etc.).



The concept of sentiment analysis has been utilized in Natural Language Processing (NLP) and in a wide range of application domains, such as education, entertainment, hosteling and other businesses \cite{medhat2014sentiment}. On the other hand, Visual sentiment analysis is relatively new and less explored. A large portion of the literature on visual sentiment/emotion recognition relies on facial expressions \cite{busso2004analysis}, where face-close up images are analyzed to predict a person's emotions. More recently, the concept of emotion recognition has been extended to relatively more complex images having multiple objects and background details. Thanks to the recent advances in deep learning, encouraging results have been recently obtained  \cite{chen2014deepsentibank,poria2018multimodal}. 


In this article, we analyze the role of visual sentiment analysis in complex disaster-related images. To the best of our knowledge, no prior work analyzes disaster-related imagery from this prospective. We also identify the challenges and potential applications with the objective of setting a benchmark for future research on visual sentiment analysis. 

The main contributions of this work can be summarized as follows:

\begin{itemize}
    \item We extend the concept of visual sentiment analysis to disaster-related visual contents, and identify the associated challenges and potential applications.
       \item In order to analyze human's perception and sentiments about disasters, we conducted a crowd-sourcing study to obtain annotations for the experimental evaluation of the proposed visual sentiment analyzer.
            \item We propose a multi-label classification framework for sentiment analysis, which also helps in analyzing the correlation among sentiments/tags.
               \item Finally, we conduct experiments on a newly collected dataset to evaluate the performance of the proposed visual sentiment analyzer. 
\end{itemize}

The rest of the paper is organized as follows: Section 2 provides detailed description of the related work; Section 3 describes the proposed methodology; Section 4 provides detailed description of the experimental setup, conducted experiments, and detailed analysis of the experimental results; Section 5 provides concluding remarks and identifies directions of future research. 

 

\section{Related Work}

In contrast to other research domains, such as NLP, the concept of sentiment analysis is relatively new in visual content analysis. The research community has demonstrated an increasing interest in the topic and a variety of techniques have been proposed with particular focus on the feature extraction and classification strategies. The vast majority of the efforts in this regard aim to analyze and classify face-closeup images for different types of sentiments/emotions and expressions. Busso et al. \cite{busso2004analysis} rely on facial expressions along with speech and other information in a multimodal framework. Several experiments have been conducted to analyze and compare the performance of different sources of information, individually and in different combination, in support of human emotions/sentiment recognition. A multimodal information based approach has also been proposed in \cite{poria2018multimodal}, where facial expressions are jointly utilized with textual and audio features that are extracted from videos. Facial expressions are extracted through the Luxand FSDK 1.7\footnote{https://www.luxand.com/facesdk/} open source library along with GAVAM features \cite{saragih2009face}. Textual and audio features are extracted through the Sentic computing paradigm \cite{cambria2010sentic} and OpenEAR \cite{eyben2009openear}, respectively. Next, different feature and decision-level fusion methods are used to jointly exploit the visual, audio, and textual information for the task.

More recently, the concept of emotion/sentiment analysis has been extended to more complex images involving multiple objects and background details \cite{kim2018building,chen2014deepsentibank,wang2016beyond,constantin2019computational}. For instance, Wang et al. \cite{wang2015unsupervised} rely on mid and low-level visual features along with textual information for sentiment analysis in social media images. Chen et al. \cite{chen2014deepsentibank} proposed DeepSentiBank, a deep convolutional neural network-based framework for sentiment analysis of social media images. To train the proposed deep model, around one million images with strong emotions have been collected from Flickr. In \cite{wang2016beyond}, Deep Coupled Adjective and Noun neural networks (DCAN), is proposed for sentiment analysis without the traditional Adjective Noun Pairs (ANP) labels. The framework is composed of three different networks, each aiming to solve a particular challenge associated with sentiment analysis. Some methods also utilized existing pre-trained models for sentiment analysis. For instance, Campose et al. \cite{campos2015diving} fine-tuned CaffeNet \cite{jia2014caffe}, on a newly collected dataset for sentiment analysis conducting experiments to analyze the relevance of the features extracted through different layers of the network. In \cite{peng2015mixed} existing pre-trained CNN models are fine-tuned on a self-collected dataset. The dataset contains images from social media, which are annotated through a crowd-sourcing activity involving human annotators. Kim et al. \cite{kim2018building} also rely on the transfer learning techniques for their proposed emotional machine. Object and scene-level information, extracted through deep models pre-trained on ImageNet and Places datasets, respectively, have been jointly utilized for this purpose. Color features have also been employed to perceive the underlying emotions. 

\section{Proposed Methodology}
\begin{figure*}[ht!]
\centering
\includegraphics[width=0.96\linewidth]{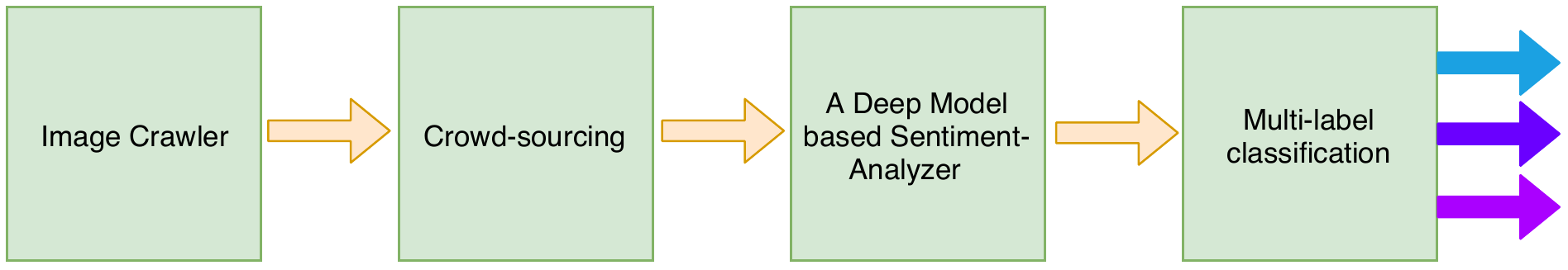}
\caption{Block diagram of the proposed framework for visual sentiment analysis.}
	\label{methodology}
\end{figure*}

Figure \ref{methodology} provides the block diagram of the framework implemented for visual sentiment analysis. As a first step, social media platforms are crawled for disaster-related images using different keywords (floods, hurricanes, wildfires, droughts, landslides, earthquakes, etc.). The downloaded images are filtered manually and a selected subset of images are considered for the crowd-sourcing study in the second step where a large number of participants tagged the images. A CNN and a transfer learning method is used for multi-label classification to automatically assign sentiments/tags to images. In the next subsections, we provide a detailed description of the crowd-sourcing activity and the proposed visual deep sentiment analyzer.

\subsection{The crowd-sourcing study}
\begin{figure*}[ht!]
\centering
\includegraphics[width=0.9\linewidth]{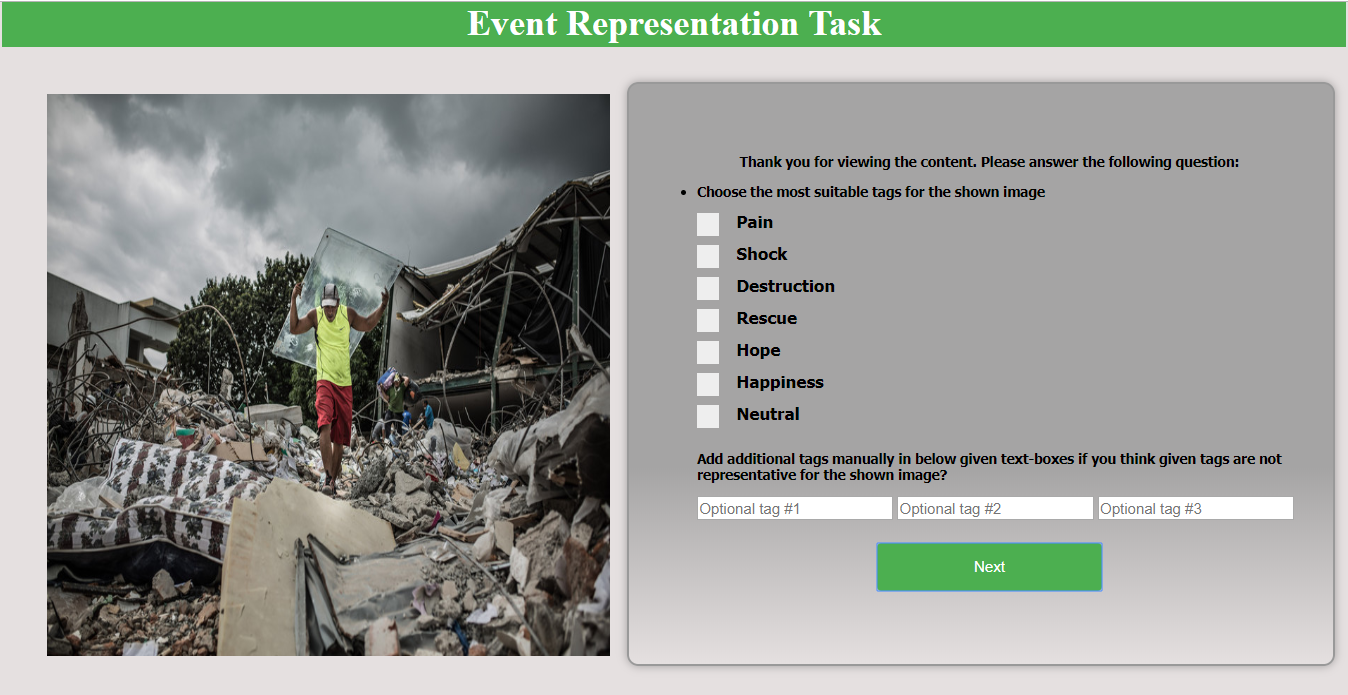}
\caption{Illustration of the platform used for the crowd-sourcing study. A disaster-related image and several tags are presented to the users for association. The users' are also encouraged to provide additional tags.}
	\label{crowdsourcing_study}
\end{figure*}

In order to analyze human's perception and sentiments about disasters and how they perceive disaster-related images, we conducted a crowd-sourcing study. The study is carried out online through a web application specifically developed for the task, which was shared with participants including students from University of Trento (Italy), and UET Peshawar (Pakistan) as well as with other contacts with no scientific background. Figure \ref{crowdsourcing_study} provides an illustration of the platform we used for the crowd-sourcing study. In the study, participants were provided with a disaster-related image, randomly selected from the pool of images, along with a set of associated tags. The participants were then asked to assign a number of suitable tags, which they felt relevant to the image. The participants were also encouraged to associate additional tags to the images, in case they felt that the provided tags were not relevant to the image. 

One of the main challenges in the crowd-sourcing study was the selection of the tags/sentiments to be provided to the users. In the literature, sentiments are generally represented as \textit{Positive}, \textit{Negative} and \textit{Neutral} \cite{medhat2014sentiment}. However, considering the specific domain we are addressing (natural and man-made disasters) and the potential applications of the proposed system, we are also interested in tags/sentiments that are more specific to adverse events, such as pain, shock, and destruction, in addition to the three common tags. Consequently, we opted for a data-driven approach, by analyzing users' tags associated with disaster images crawled form social media outlets. Apart from the sentimental tags, such as \textit{pain}, \textit{shock} and \textit{hope}, we also included some additional tags, such as \textit{rescue} and \textit{destruction}, which are closely associated with disasters and can be useful in different applications utilized by online news agencies, humanitarian, and non-governmental organizations (NGOs). The option for adding additional tags also helps to take the participants' viewpoints into account. 

The crowd-sourcing activity was carried out on 400 images related to 6 different types of disasters: earthquakes, floods, droughts, landslides, thunderstorms, and wildfires. In total, we obtained 2,587 responses from the users, with an average of 6 users per image. We made sure to have at least 5 different users for each image. Table \ref{tab:statistics_cs} provides the statistics of the crowd-sourcing study in terms of the total number of times each tag has been associated with images by the participants. As can be seen in Table \ref{tab:statistics_cs}, some tags, such as \textit{destruction}, \textit{rescue} and \textit{pain}, are used more frequently compared to others. 
\begin{table}[]
\caption{Statistics of the crowd-sourcing study in terms of of the total number of times each tags has been associated with images by the participants.}
\label{tab:statistics_cs}
\begin{center}
\begin{tabular}{|c|c|}
\hline
\textbf{Sentiments/tags} & \textbf{Count} \\ \hline
Destruction & 871 \\ \hline
Happiness & 145 \\ \hline
Hope & 353 \\ \hline
Neutral & 214 \\ \hline
Pain & 454 \\ \hline
Rescue & 694 \\ \hline
Shock & 354 \\ \hline
\end{tabular}
\end{center}
\end{table}

\begin{figure*}[ht!]
\centering
\includegraphics[width=0.9\linewidth]{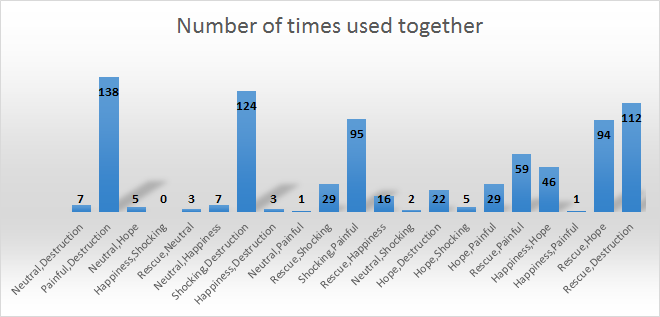}
\caption{Correlation of tag pairs: number of times different tag pairs used by the participants of the crowd-sourcing study to describe the same image.}
	\label{correlation_tags}
\end{figure*}

During the analysis of the responses from the participants, we observed that certain tag pairs have been used to describe images. For instance, pain and destruction, hope and rescue, shock and pain, are used several times jointly. Similarly, shock, destruction and pain have been used jointly 59 times. The three tags: rescue, hope, and happiness, are also used often together. This correlation among the tag/sentiment pairs provides the foundation for our multi-label classification, as opposed to single-label multi-class classification, of the sentiments associated with disasters-related images. Figure \ref{correlation_tags} shows the number of times the sentiments/tags are used together by the participants in the crowd-sourcing activity. For final annotation, the decision is made on the basis of majority votes from the participants of the crowd-sourcing study.

\subsection{The Visual Sentiment Analyzer}

The proposed framework for visual sentiment analysis is inspired by the multi-label image classification framework \footnote{https://towardsdatascience.com/multi-label-image-classification-with-inception-net-cbb2ee538e30} and is mainly based on a Convolutional Neural Network (CNN) and a transfer learning method, where the model pre-trained on ImageNet is fine-tuned for visual sentiment analysis. In this work, we analyze the performance of several deep models such as AlexNet \cite{krizhevsky2012imagenet}, VggNet \cite{simonyan2014very}, ResNet \cite{he2016deep} and Inception v-3 \cite{szegedy2016rethinking} as potential alternatives to be employed in the proposed visual sentiment analysis framework.

The multi-label classification strategy, which assigns multiple labels to an image, better suits our visual sentiment classification problem and is intended to show the correlation of different sentiments. In order for the network to fit the task of visual sentiment analysis, we introduced several changes to the model as will be described in the next paragraph.

\subsection{Experimental Setup}

In order to fit the pre-trained model to multi-label classification, we create a ground truth vector containing all the labels associated with an image. We also made some modifications in the existing pre-trained Inception-v3 \cite{szegedy2016rethinking} model by extending the classification layer to support multi-label classification. To do so, we replaced the soft-max function, which is suitable for single-label multi-class classification, and squashes the values of a vector into a  \textit{[0,1]} range holding the total probability, with a sigmoid function. The motivation for using a sigmoid function comes from the nature of the problem, where we are interested to express the results in probabilistic terms; for instance, an image belongs to the class \textit{shock} with 80\% probability and to class \textit{destruction} and \textit{pain} with 40\% probability. Moreover, in order to train the multi-label model properly, the formulation of the cross entropy is also modified accordingly (i.e., replacing softmax with sigmoid function). For the multiple labels, we modify the top layer to obtain posterior probabilities for each type of sentiment associated with an underlying image. 

The dataset used for our experimental studies has been divided into training (60\%), validation (10\%), and evaluation (30\%) sets.

\section{Experiments and Evaluations}
The basic motivation behind the experiments to provide a baseline for the future work in the domain. To this aim, we evaluate the proposed multi-label framework for visual sentiment analysis using several existing pre-trained state-of-the-art deep learning models including: AlexNet, VggNet, ResNet, and Inception v3. Table \ref{tab:results_60} provides the experimental results obtained using these deep models.

\begin{table}[]
\caption{Evaluation of the proposed visual sentiment analyzer with different deep learning models pre-trained on ImageNet.}
\label{tab:results_60}
\begin{tabular}{|c|c|}
\hline
Model & Accuracy (\%) \\ \hline
AlexNet & 79.69 \\ \hline
VggNet & 79.58 \\ \hline
Inception-v3 & 80.70 \\ \hline
ResNet & 78.01 \\ \hline
\end{tabular}
\end{table}

Considering the complexity of the task and the limited amount of training data, the obtained results are encouraging. Though there's no significant difference in the performance of the models, slightly better results are obtained with Inception-v3 models. Lowest accuracy has been observed for ResNet, but such reduction in the performance could be due to the size of the dataset used for the study.  

\begin{figure}
  \begin{subfigure}{}
  \includegraphics[height=0.20\linewidth,width=0.20\linewidth]{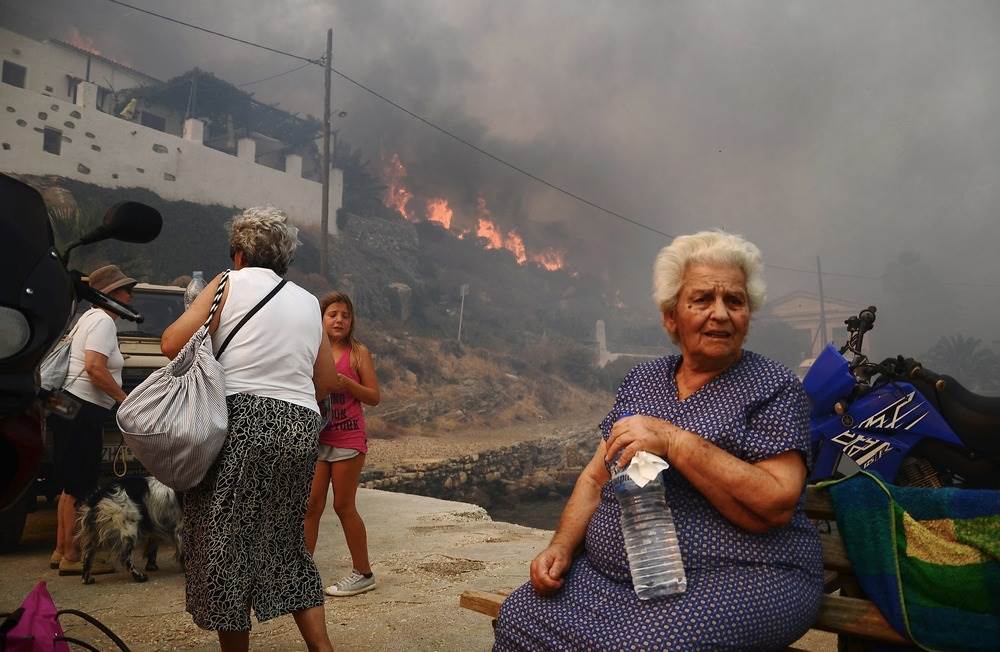}\hfill
  \includegraphics[height=0.20\linewidth,width=0.20\linewidth]{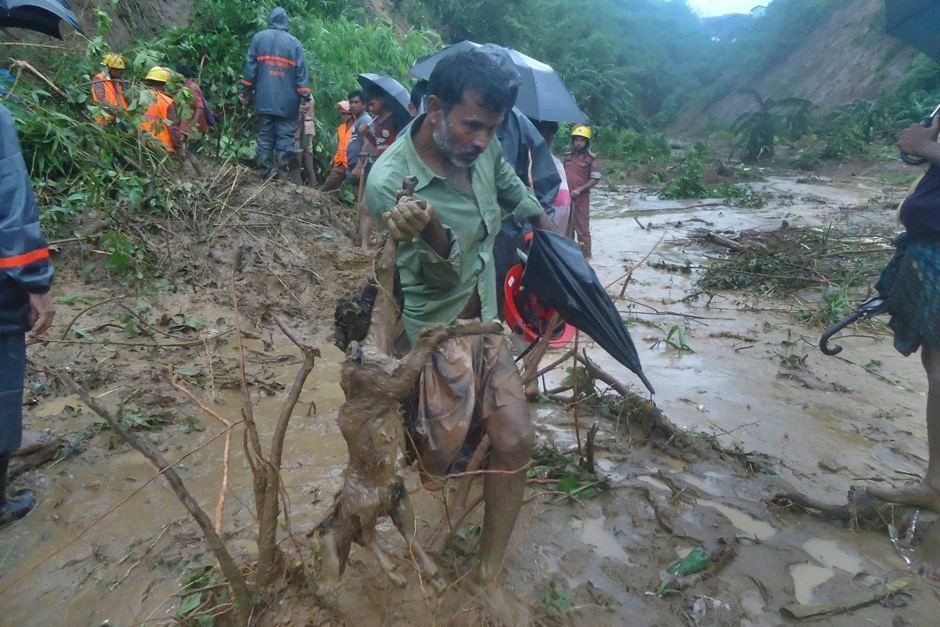}\hfill
  \includegraphics[height=0.20\linewidth,width=0.20\linewidth]{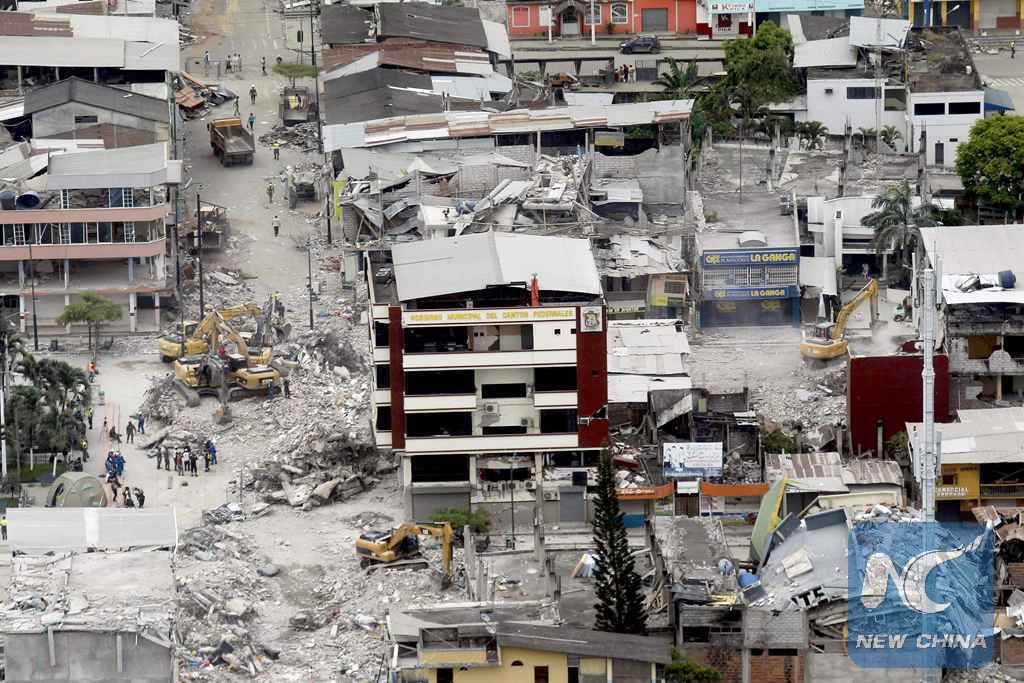}\hfill
  \includegraphics[height=0.20\linewidth,width=0.20\linewidth]{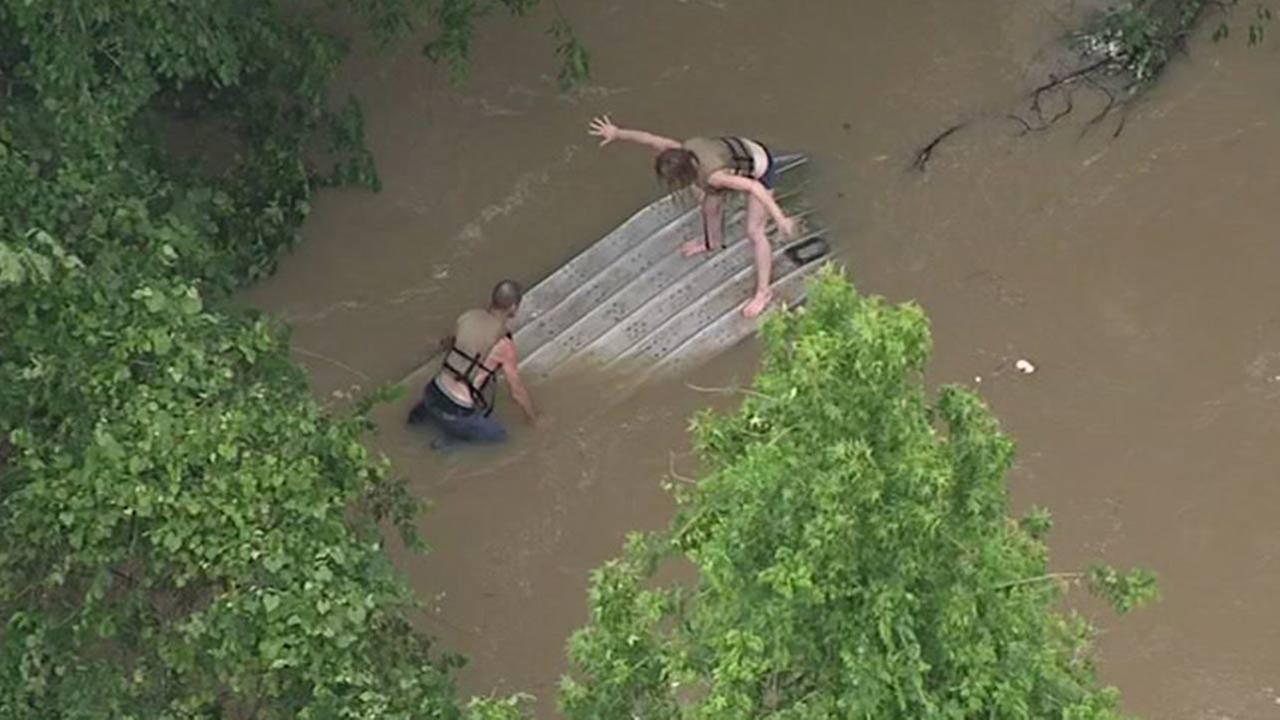}\hfill
  \end{subfigure}
  \caption{Some sample output of the proposed visual sentiment analyzer.}
  \label{fig:sample_output}
\end{figure}

In order to show the effectiveness of the proposed visual sentiment analyzer, we also provide some sample output images in Figure \ref{fig:sample_output}, showing the output of the proposed visual sentiment analyzer in terms of the percentage/probabilities for each label. Table \ref{tab:results_60} provides the statistics for these samples in terms of the probability for each label and probabilities/percentages computed through human annotators. Due to space limitation, only four samples are provided in the paper to give an idea about the performance of the method. For this particular qualitative analysis, we converted the responses of the participants of the crowd sourcing study into percentages (i.e., the degree to which each image belongs to a particular label) for each label associated with each image. These percentages are different from the ground truth used during training and evaluation where images were assigned labels on a majority voting basis. For instance, the percentages based on the responses of the crowd sourcing study for the first image (leftmost in Figure \ref{fig:sample_output}) are: \textit{destruction} =0.10, \textit{happiness} =0.0, \textit{hope} =0.10, \textit{neutral} = 0.0, \textit{pain} =0.35, \textit{rescue} = 0.30 and \textit{shock} = 0.20 while the output of the proposed visual sentiment analyzer in terms of probabilities for each label/class are: \textit{destruction} =0.16, \textit{happiness} =0.04, \textit{hope} =0.06, \textit{neutral} = 0.02, \textit{pain} =0.58, \textit{rescue} = 0.28 and \textit{shock} = 0.17.  In most of the cases, the proposed model provides results that are similar to the percentages obtained from the users' responses, demonstrating the effectiveness of the proposed method.

\begin{table}[]
\caption{Sample outputs in terms of ground truth obtained from users in terms of percentage in the crowd-sourcing study vis-a-vis predicted probabilities.}
\label{tab:results_60}
\begin{tabular}{|c|l|c|l|c|c|l|l|c|c|l|c|l|c|l|}
\hline
\multirow{2}{*}{\textbf{Image}} & \multicolumn{2}{l|}{\textbf{Destruction}} & \multicolumn{2}{l|}{\textbf{Happiness}} & \multicolumn{2}{c|}{\textbf{Hope}} & \multicolumn{2}{l|}{\textbf{Neutral}} & \multicolumn{2}{c|}{\textbf{Pain}} & \multicolumn{2}{c|}{\textbf{Rescue}} & \multicolumn{2}{c|}{\textbf{Shock}} \\ \cline{2-15} 
 & GT & Pred. & GT & Pred. & GT & Pred. & GT & Pred. & GT & Pred. & GT & Pred. & GT & Pred. \\ \hline
1 & 0.10 & 0.16 & 0.0 & 0.04 & 0.1& 0.06 & 0 & 0.027 & 0.35& 0.58 & 0.30& 0.28 & 0.20 &0.17   \\ \hline
2 & 0.24 & 0.24 & 0.0 & 0.05 & 0.0 & 0.08 & 0.34 & 0.36 & 0.429 & 0.44 & 0.514 & 0.59 & 0.20 & 0.33 \\ \hline
3 & 0.167 & 0.23 & 0.0 & 0.05 & 0.10 & 0.13 & 0.16 & 0.17 & 0.46 & 0.59 & 0.33 & 0.26 & 0.0 & 0.13 \\ \hline
4 & 0.10 & 0.18 & 0.0 & 0.03 & 0.09 & 0.05 & 0.20 & 0.26 & 0.0 & 0.33 & 0.72 & 0.72 & 0.0 & 0.20 \\ \hline
\end{tabular}
\end{table}

\section{Conclusions, challenges and Future work}

In this paper, we addressed the challenging problem of visual sentiment analysis of disaster-related images obtained from social media. We analyzed how people respond to disasters and obtained their opinions, attitudes, feelings, and emotions toward the disaster-related images through a crowd-sourcing activity. We show that the visual sentiment analysis/emotions recognition, though a challenging task, can be carried out on more complex images using some deep learning techniques. We also identified the challenges and potential applications of this relatively new concept, which is intended to set a benchmark for future research in visual sentiment analysis. 

Though the experimental results obtained during the initial experiments on the limited dataset are encouraging, the task is challenging and needs to be investigated in more details. Specifically, the reduced availability of suitable training and testing images is probably the biggest limitation. Since visual sentiment analysis aims to present human's perception of an entity, crowd-sourcing seems to be a valuable option to acquire training data for automatic analysis. 
In terms of visual features, we believe that object and scene-level features can play complementary roles in representing the images. Moreover, multi-modal analysis will further enhance the performances of the proposed sentiment analyzer. 
This suggests that within the domain of purely visual information, the conveyed information can differ, suggesting that the interpretation of the image is subject to change depending on the level of detail, the visual perspective, and the intensity of colors. 
We expect these elements to play a major role in the evolution of frameworks like the one we have presented, and when combined with additional media sources (e.g., audio, text, meta-data), can provide a well rounded perspective about the sentiments associated with a given event. 
%
%
%
 \bibliographystyle{splncs04}
 \bibliography{mybibliography}
\end{document}